\documentclass[a4paper,conference]{IEEEtran}
\usepackage[left=0.62in,right=0.62in,top=0.38in,bottom=1in]{geometry}
\usepackage{cite}
\usepackage{amsmath,amssymb,amsfonts}
\usepackage{algorithmic}
\usepackage{graphicx}
\usepackage{textcomp}
\usepackage{xcolor}
\usepackage{url}

\begin{document}

\title{Non-Spatial Hash Chemistry as a Minimalistic Open-Ended Evolutionary System}

\author{\IEEEauthorblockN{Hiroki Sayama}
\IEEEauthorblockA{\textit{Binghamton Center of Complex Systems}\\ 
\textit{Binghamton University, State University of New York}\\
Binghamton, NY, USA \\
\textit{Waseda Innovation Lab}\\
\textit{Waseda University}\\
Shinjuku, Tokyo, Japan\\
sayama@binghamton.edu}}


\maketitle

\begin{abstract}
There is an increasing level of interest in open-endedness in the recent literature of Artificial Life and Artificial Intelligence. We previously proposed the cardinality leap of possibility spaces as a promising mechanism to facilitate open-endedness in artificial evolutionary systems, and demonstrated its effectiveness using Hash Chemistry, an artificial chemistry model that used a hash function as a universal fitness evaluator. However, the spatial nature of Hash Chemistry came with extensive computational costs involved in its simulation, and the particle density limit imposed to prevent explosion of computational costs prevented unbounded growth in complexity of higher-order entities. To address these limitations, here we propose a simpler non-spatial variant of Hash Chemistry in which spatial proximity of particles are represented explicitly in the form of multisets. This model modification achieved a significant reduction of computational costs in simulating the model. Results of numerical simulations showed much more significant unbounded growth in both maximal and average sizes of replicating higher-order entities than the original model, demonstrating the effectiveness of this non-spatial model as a minimalistic example of open-ended evolutionary systems.
\end{abstract}


\section{Introduction}

The {\em open-endedness} of learning and evolving artificial systems has become a major topic of interest in the recent literature of Artificial Life (AL) and Artificial Intelligence (AI) \cite{lehman2008exploiting,sayama2011seeking,taylor2016open,
banzhaf2016defining,taylor2018routes,sayama2018seeking,stanley2019open,
packard2019overview,stepney2021modelling,borg2023evolved,stepney2023open,
samvelyan2024tweet}. In this subcommunity of AL/AI, researchers have been trying to design a computational process that keeps exploring a vast possibility space and continuously discovers and generates various novel solutions indefinitely, rather than converging to an optimal solution at the fastest rate. Such open-ended nature is often considered as a fundamental essence of living systems and other dynamically sustainable systems \cite{ruiz2004universal,sayama2019suppleness,fisher2023} and may also be a product of evolution in itself \cite{szathmary2015toward,pattee2019evolved}.

As a promising mechanism to facilitate open-endedness in artificial evolutionary systems, we previously argued the importance of a ``cardinality leap'' of possibility spaces \cite{sayama2019} and demonstrated its effectiveness using Hash Chemistry, a simple artificial chemistry \cite{dittrich2001artificial,banzhaf2015artificial} model that used a hash function as a universal fitness evaluator \cite{sayama2019}. In Hash Chemistry, individual particles of various elements distribute in a two-dimensional finite space, and a group of spatially proximate particles are randomly selected and their replication/survival/death is determined based on the hash value of the group of those particles altogether. This seemingly extremely simple model nonetheless exhibited a spontaneous increase of the complexity (size) of replicating higher-order entities and the number of novel higher-order entity types. However, the spatial nature of the original Hash Chemistry model came with extensive computational costs involved in its simulation, as pointed out as one of the major limitations in \cite{sayama2019}. Moreover, the particle density limit imposed to prevent explosion of computational costs prevented unbounded growth in complexity of higher-order entities. 

To address these limitations, here we propose a simpler non-spatial variant of Hash Chemistry in which spatial proximity of particles are represented explicitly in the form of multisets of individual entities. This model modification achieved a significant reduction of computational costs in simulating the model, allowing for a large number of numerical simulations to be done within a realistic amount of time. Numerical experiments showed that this non-spatial Hash Chemistry model exhibits much more significant unbounded complexity growth of replicating higher-order entities than the original model. This paper reports the details of the new non-spatial Hash Chemistry model and the results of extensive numerical simulations.

\section{Original Model: Hash Chemistry}

\begin{figure*}[t!]
\centering
\includegraphics[width=\textwidth]{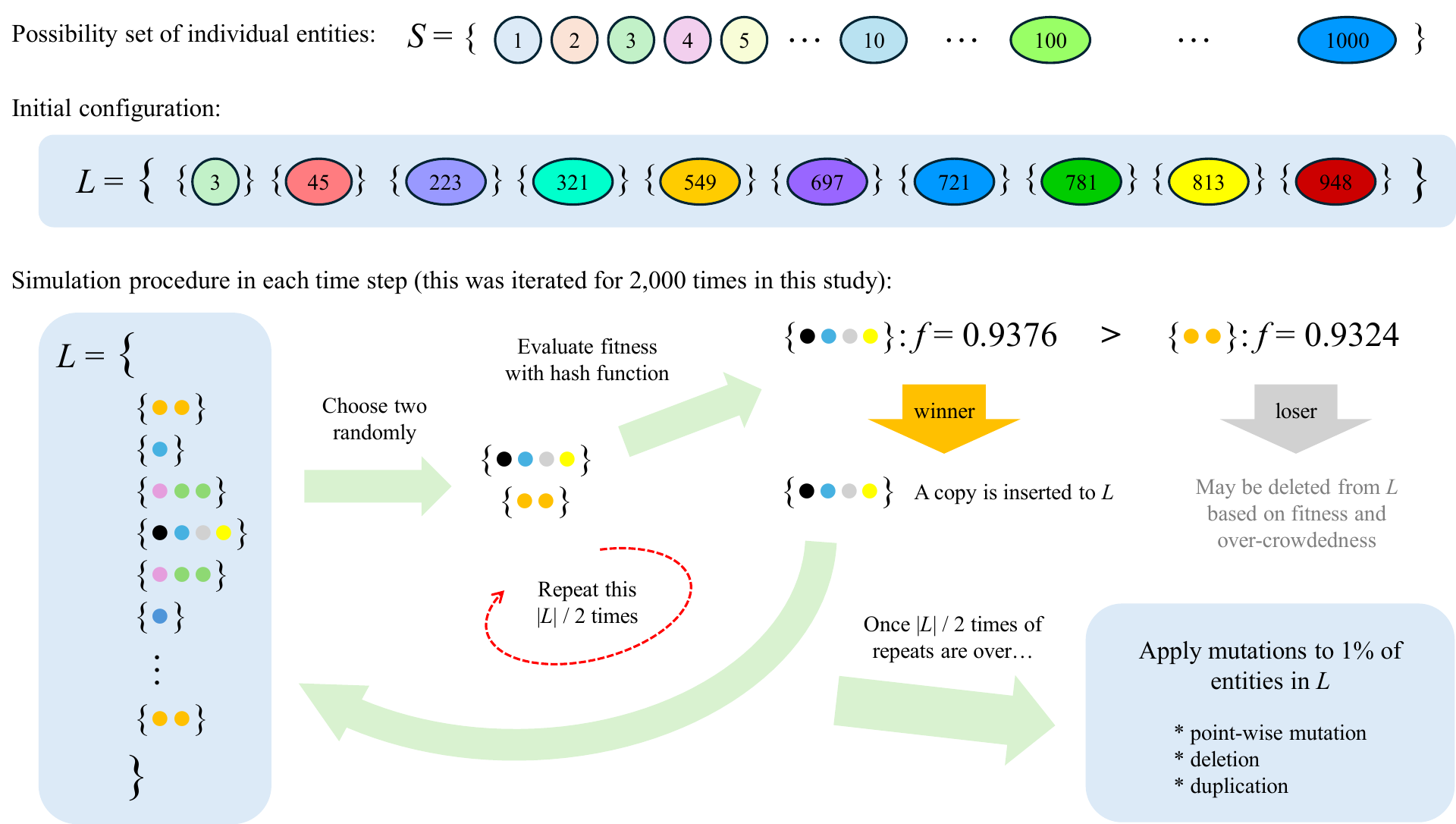}
\caption{Schematic illustration of the outline of the proposed non-spatial Hash Chemistry model. See the main text for details.}
\label{fig:outline}
\end{figure*}

Our proposed non-spatial Hash Chemistry model is largely based on the original Hash Chemistry \cite{sayama2019}. The simulation settings and algorithm of the original model are summarized as follows (taken and modified from \cite{sayama2019}):
\begin{itemize}
\item The possibility set $S$ of individual entities (types) are natural numbers ranging from 1 to 1,000, i.e., $S = \{1, 2, \ldots, 1000\}$.
\item The space is a two-dimensional continuous unit square with cut-off boundaries.
\item The initial configuration is made of 10 individual entities of randomly chosen types, randomly distributed within the space.
\item Each simulation is run for 2,000 iterations (time steps).
\item Each iteration consists of the following steps:
\begin{enumerate}
\item Move each of the individual entities a little randomly.
\item For each of the positions of individual entities, do the following:
\begin{enumerate}
\item Create a set $N$ of individual entities that are of close distance from the focal individual entity.
\item Choose a random subset $s$ of $N$ by randomly selecting $k$ entries from $N$, where $k$ is a random integer in $\{1, 2, \ldots, |N|\}$. 
\item With probability $1/|s|$, do the following:
\begin{enumerate}
\item Create a sorted list of types of the individual entities in $s$.
\item Calculate the fitness $f$ of $s$ by applying a hash function to the above list. The output is mapped to a $[0,1)$ fitness range by computing $(h \; \mathrm{mod} \; m) / m$, where $h$ is the hash value (Mathematica's ``Hash'' function and $m=100,000$ were used in \cite{sayama2019}).
\item With probability $1-f$, delete all individual entities in $s$ from the space. {\em (death)}
\item With probability $f (1-|N|/d_\mathrm{max})$, where $d_\mathrm{max}$ is the maximum density of entities ($d_\mathrm{max} = 100$ in \cite{sayama2019}), add copies of all individual entities in $s$ to the space. {\em (replication)}
\end{enumerate}
\end{enumerate}
\item For each of the individual entities, change its type to a type randomly sampled from $S$ with a small probability. {\em (mutation)}
\item Randomize the order of individual entities.
\end{enumerate}
\end{itemize}

The spatial nature of the original Hash Chemistry model resulted in extensive computational costs, especially in Steps 2-a (neighbor detection) and 2-c-iii (deletion of individual entities) in the above simulation algorithm. In particular, Step 2-a's worst-case computational complexity is $O(n^2)$, where $n$ is the total number of individual entities in the environment. This can easily become very expensive when $n$ grows very large.

\section{Proposed Model: Non-Spatial Hash Chemistry}
\label{algorithm}

To address the issue of computational costs, we simplified the original model by removing the spatial extension and representing clusters of proximate individual entities explicitly in the form of multisets. More specifically, we treated multisets of individual entity types as the basic unit of replication (analogous to molecules) and defined the entire evolutionary system simply as a well-mixed population of such multisets, without any explicit spatial coordinates involved. This modification eliminated the need for neighbor detection in space and thus sped up numerical simulation substantially. 

The evolution of replicating entities is simulated via repeated pairwise competitions between two randomly chosen multisets. The details of the simulation settings and algorithm of the revised model are as follows (also see Fig.\ \ref{fig:outline} for schematic illustration of the model outline):
\begin{itemize}
\item The possibility set $S$ of individual entities (types) are natural numbers ranging from 1 to 1,000, i.e., $S = \{1, 2, \ldots, 1000\}$.
\item There is no spatial domain assumed; the system configuration is represented simply by a list of multisets of individual entities $L$.
\item The initial configuration of $L$ is a list that contains 10 singleton sets, each of which contains an individual entity of a randomly chosen type.
\item Each simulation is run for 2,000 iterations (time steps).
\item Each iteration consists of the following steps:
\begin{enumerate}
\item Repeat the following $n/2$ times\footnote{This number was set such that each multiset would be involved in a competition for about once in each time step, on average.} where $n=|L|$:
\begin{enumerate}
\item Choose two multisets from $L$ randomly.
\item Use the hash function to calculate the fitness of the selected two multisets. The output of the hash function is mapped to a $[0,1)$ fitness range by computing $f = (h \; \mathrm{mod} \; m) / m$, where $h$ is the hash value and $f$ is the calculated fitness. Mathematica's ``Hash'' function and $m=100,000,000$ were used in this study\footnote{We used a much larger $m$ compared to \cite{sayama2019} to increase the fitness values' resolution near their maximal value 1.}.
\item Insert a copy of the multiset with greater $f$ to $L$. {\em (replication)}
\item Remove the multiset with smaller $f$ from $L$ with probability $1 - f (1-n/n_\mathrm{max})$, where $n_\mathrm{max}$ is the carrying capacity of the environment ($n_\mathrm{max} = 10,000$ in this study). {\em (death)}
\end{enumerate}
\item Apply mutations to 1\% of all the multisets in $L$. When mutations are applied to a multiset, each individual entity in the multiset is subject to change with 20\% probability. 80\% of such point-wise changes are random entity type swaps within $S$, while the remaining 20\% are removals of the individual entity from the multiset. Also, with 20\% probability, the multiset's content is duplicated. If mutations result in an empty multiset, it is removed from $L$. {\em (mutation)}
\end{enumerate}
\end{itemize}

\section{Performance Comparison}

We implemented both the original and proposed non-spatial Hash Chemistry models in Mathematica 14.0\footnote{Codes are available from the author upon request.} and conducted numerical simulations from randomly generated initial conditions to compare computational performance of the two models. For this purpose, we removed visualization and data logging functionalities from the codes and just focused on the model simulation only. To do a fair performance comparison, we used Mathematica's ``Nearest'' function in its optimized settings for nearest neighbor detection in the original model. We  used a Windows 11 (64-bit) desktop workstation with an Intel i9 CPU (10 cores) at 3.70 GHz with 64 GB RAM for all the simulations.

\begin{figure}
\centering
\includegraphics[width=\columnwidth]{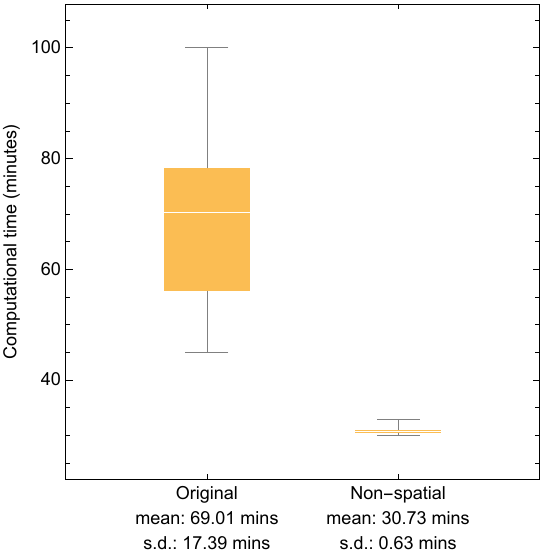}
\caption{Box-whisker plots comparing the distributions of computational time needed to complete one simulation run for 2,000 iterations between the original Hash Chemistry model \cite{sayama2019} (left) and the non-spatial version proposed in this paper (right). The vertical axis shows the length of simulation time in minutes on a Windows 11 (64-bit) desktop workstation with an Intel i9 CPU (10 cores) at 3.70 GHz with 64 GB RAM. The original model often exhibited extinction of particles in the very early stage of simulation, and hence those trivial extinction cases were excluded from the distribution of the original model. No such extinction occurred in the non-spatial model. Overall, the non-spatial model achieved a 69.01 / 30.73 = 2.25 times speed-up compared to the original model.}
\label{fig:performance-comparison}
\end{figure}

We conducted 20 independent simulation runs for performance measurement of each of the two versions of Hash Chemistry on the above same hardware. In the original model, there were 3 out of 20 cases of premature extinction of replicating entities, which were excluded from data visualization and analysis. Figure \ref{fig:performance-comparison} compares the computational time needed to complete one simulation run for 2,000 iterations in the two models. Although a rigorous direct performance comparison would not be possible because the two models were essentially simulating fundamentally different dynamical processes with different numbers of components\footnote{The original model typically involved only a few thousands of individual particles, whereas the new non-spatial model involved as many as 10,000 multisets each of which could contain multiple individual particles.}, the practical performance difference was quite clear in Fig.\ \ref{fig:performance-comparison}. In the original model, each simulation run took nearly 70 minutes on average\footnote{Note that this is already significantly faster than the simulations reported in \cite{sayama2019}, which were run on older hardware and involved intensive data logging but did not use optimized neighbor detection methods.}, and the variance of computational time was also very large. In contrast, in the new non-spatial version, each simulation run took just about 30 minutes on average with very little variance. Overall, our non-spatial model achieved a 69.01 / 30.73 = 2.25 times speed-up compared to the original model. 

Another notable difference is that, unlike in the original model, extinction of replicating entities never happened in the simulations of the non-spatial model, because deaths would not occur so frequently until the population size approaches the carrying capacity of the environment (see Step 1-d above).

\section{Observed Evolutionary Dynamics}

We conducted another set of 100 independent simulation runs of the non-spatial Hash Chemistry model to investigate its evolutionary dynamics in detail. During each simulation run, every single replication event was saved in a log file with detailed information about the content of the replicating multiset and its fitness value. The 100 log files generated from the simulations were post-processed and analyzed for visualizations and statistical analyses, which are reported below.

\begin{figure}
\centering
\includegraphics[width=\columnwidth]{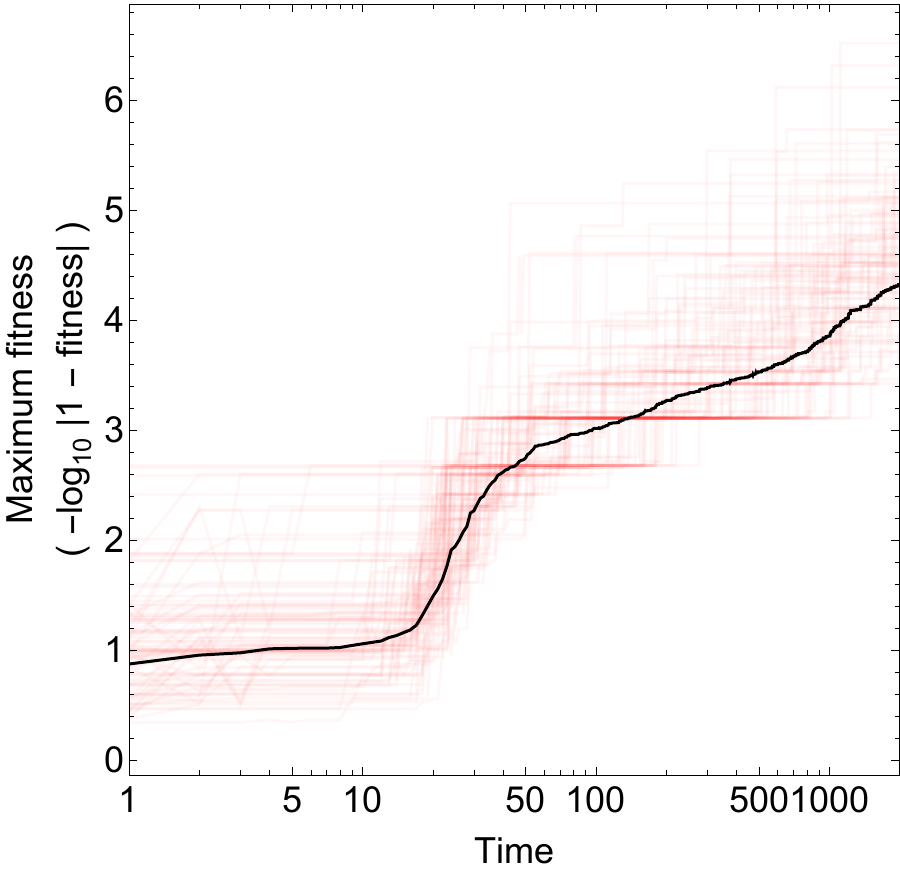}\\~\\
\includegraphics[width=\columnwidth]{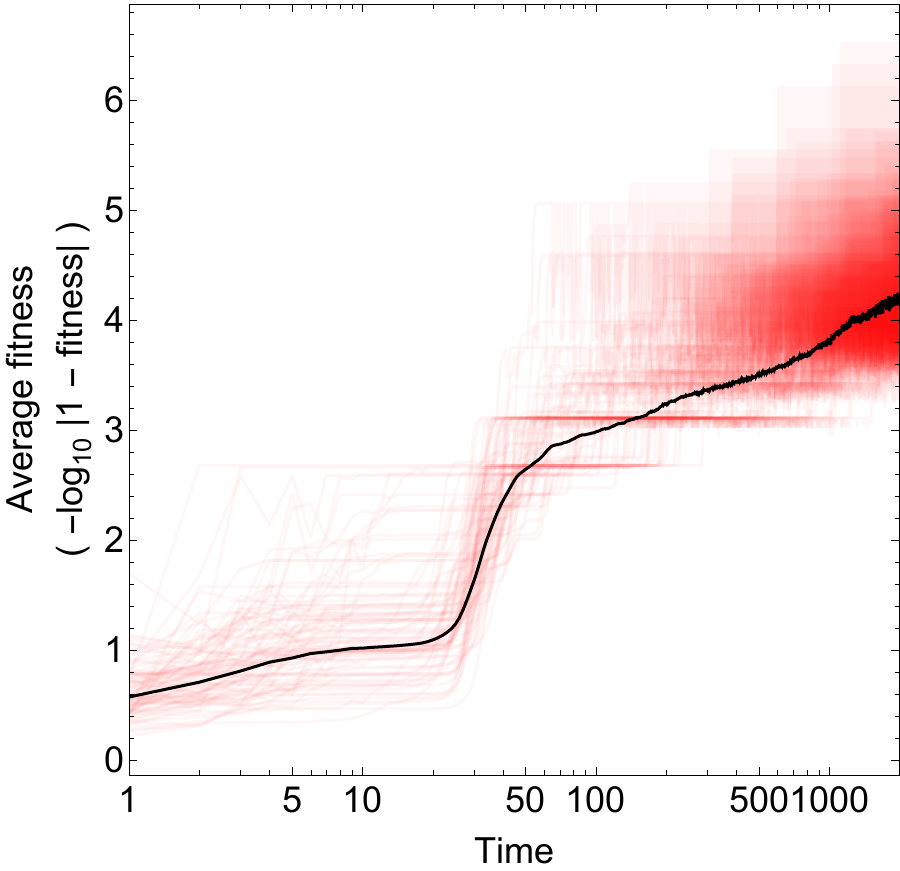}
\caption{Fitness values of successfully replicated multisets (higher-order entities) in simulations of non-spatial Hash Chemistry. Top: Maximum fitness value observed in each time step. Bottom: Average fitness value in each time step. In each plot, the red thin curves show results of 100 independent simulation runs, while the black solid curve shows their average. The time is in log scale to show long-term trends more clearly. The fitness values are visualized using $-\log_{10} | 1 - \mathrm{fitness} |$ to visualize increasingly finer improvement of the fitness that progresses over the course of simulation. In both plots, the fitness increased rapidly right before $t=50$ because the population size approached the environment's carrying capacity and the selection pressure started to kick in by then. The maximum fitness and the average fitness behaved very similarly because the population was usually dominated by a single fittest multiset type most of the time. It is also noticeable that there was a large variation among the independent simulation runs (red thin curves) because the system would easily get trapped in local fitness optima for a substantially long period of time before evolution would discover a fitter multiset.}
\label{fig:adaptation}
\end{figure}

Figure \ref{fig:adaptation} presents time series of (i) maximum fitness values and (ii) average fitness values of successfully replicated multisets (i.e., higher-order entities) observed in each time step. In these plots, the fitness values are visualized using $-\log_{10} | 1 - \mathrm{fitness} |$ (i.e., how close, or for how many digits, the fitness approaches its theoretical maximum 1), in order to visualize increasingly finer improvement of the fitness that progresses over the entire time period of simulation. The maximum/average fitness values rapidly converged toward 1 by $t=50$ when the population approaches the environment's carrying capacity and the fitness values were around 0.999 (i.e., they approached 1 for three digits). Even after that, however, the evolutionary adaptation continued for the entire simulation time period to improve the fitness further and closer for more digits to the theoretical maximum value 1. Such increasingly finer improvement of fitness was not explicitly reported in \cite{sayama2019}.

\begin{figure}
\centering
\includegraphics[width=\columnwidth]{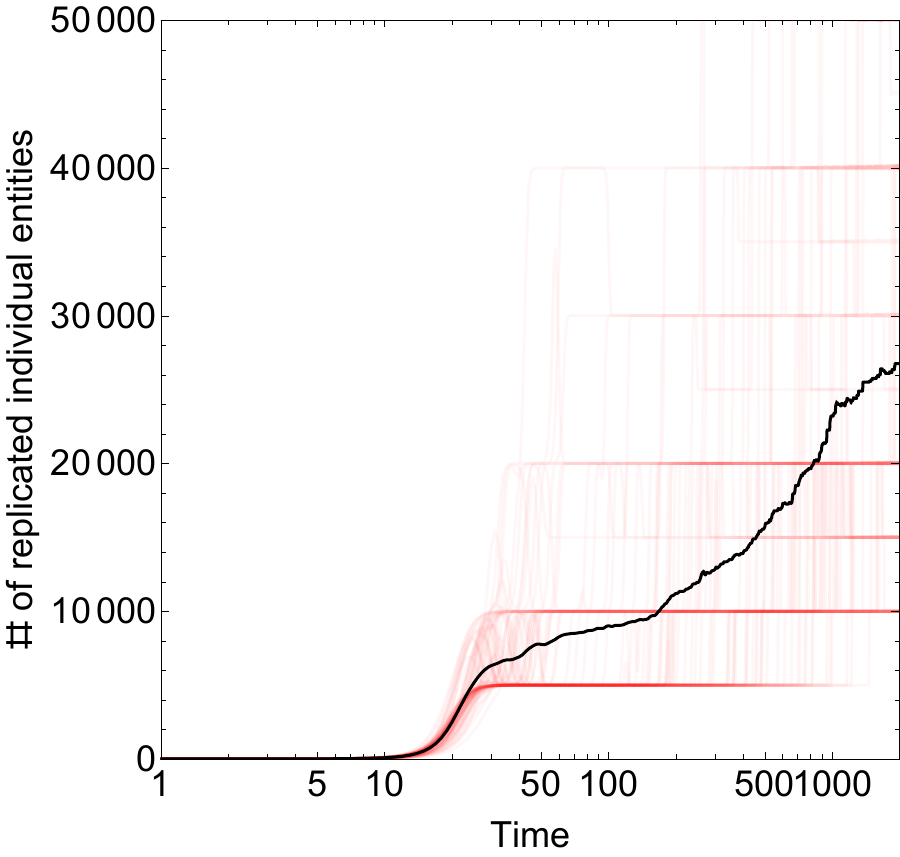}
\caption{Number of individual entities involved in replications of multisets in each time step. The red curves show results of 100 independent simulation runs, while the black solid curve shows their average. The time is in log scale to show long-term trends more clearly.}
\label{fig:adaptation2}
\end{figure}

A possible explanation of the mechanism responsible for such prolonged continuous improvement of fitness can be obtained from Figure \ref{fig:adaptation2}, which shows the number of replicated individual entities involved in replications of multisets in each time step. Since the number of pairwise matches per time step is set to half of the population size (see Step 1 above) and the environment's carrying capacity is set to 10,000 multisets (see Step 1-d above), the maximum number of replication events per time step is 5,000 in these simulations. Therefore, the increase of the total number of replicated individual entities (black solid curve) in Fig.\ \ref{fig:adaptation2} beyond 5,000 after $t \approx 25$ means that multisets containing more than one individual entity became dominant. This number kept climbing up over the entire course of simulation, implying that the evolution continued to search an increasingly larger possibility space involving a greater number of individual entities within a replicating multiset, continuously discovering multisets with increasingly (but slightly) better fitness values.

\begin{figure}
\centering
\includegraphics[width=\columnwidth]{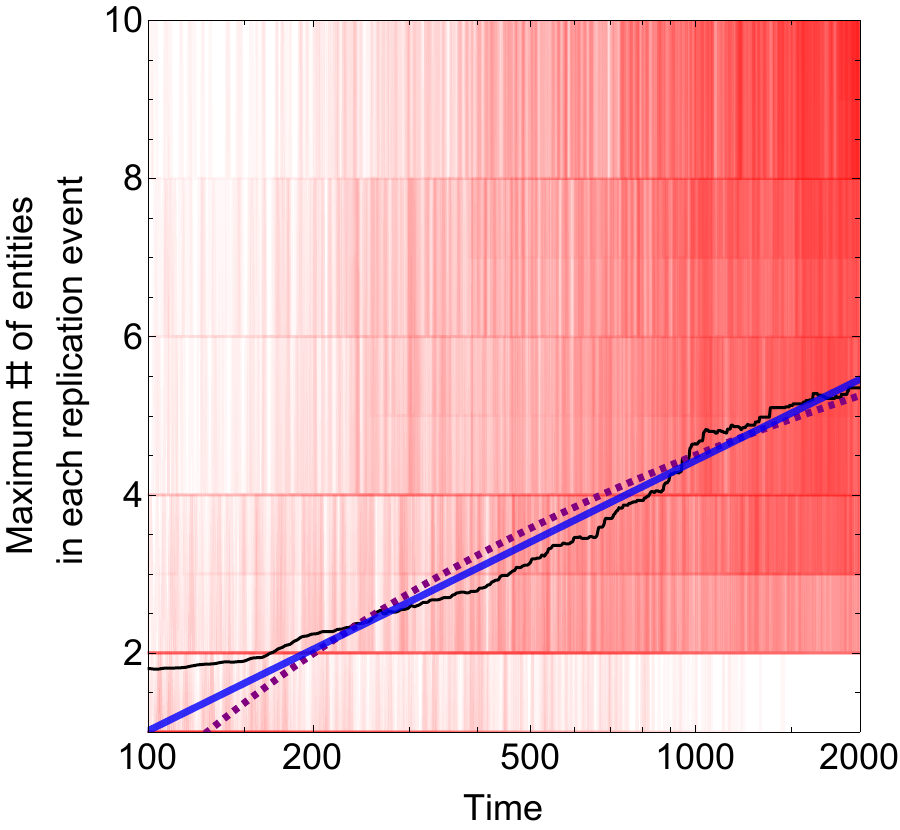}\\~\\
\includegraphics[width=\columnwidth]{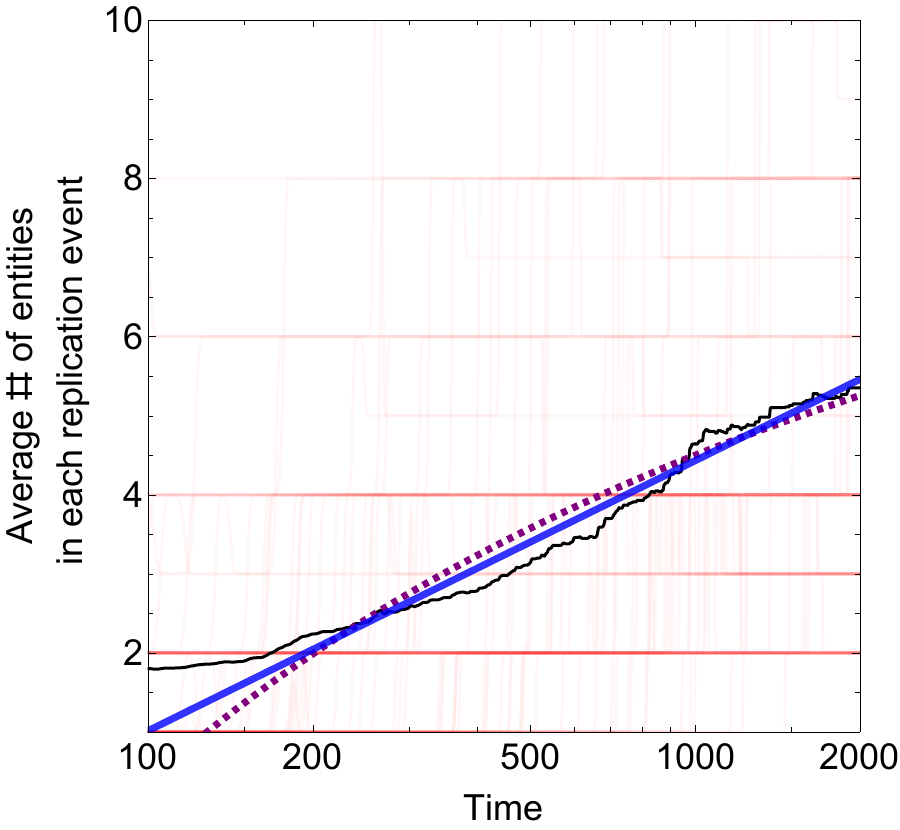}
\caption{Maximum (top) and average (bottom) numbers of individual entities in replicating multisets. The red curves show results of 100 independent simulation runs, while the black solid curve shows their average. The time is in log scale to show long-term trends more clearly. Purple (dashed) and blue (solid) curves are two different growth models (purple: bounded growth, blue: unbounded growth) fitted to the average behaviors during the time period 100--2,000. In both plots, the unbounded growth model (blue curve) was a significantly better fit. See Table \ref{tab:curvefits} for more details.}
\label{fig:higher-order}
\end{figure}

Figure \ref{fig:higher-order} shows more direct measurements of the size of replicating multisets. It was clearly observed in these plots that the dominant multisets became larger over time, demonstrating the evolutionary appearance of higher-order structures that was also confirmed in the original Hash Chemistry model \cite{sayama2019}. The fact that this key behavior, {\em ongoing growth of complexity} \cite{taylor2016open}, is also obtained in our non-spatial model means that spatial extension is not an essential factor needed to induce cardinality leap and open-ended complexity increase in an evolutionary system.

To characterize the trend of the complexity increase, two different growth models, bounded and unbounded ones, were fitted to the average behaviors during the time period 100--2,000 in logarithmic time scales, following the same analysis as in \cite{sayama2019}. Details of the models and the results are shown in Table \ref{tab:curvefits}. In both the maximum number (Fig.\ \ref{fig:higher-order} top) and the average number (Fig.\ \ref{fig:higher-order} bottom) of individual entities, the unbounded growth model was a significantly better fit to the data (see Table \ref{tab:curvefits}; compare the AIC/BIC values between bounded and unbounded growth models). This result is different from what was reported for the original model \cite{sayama2019} in which only the average number was inferred to be unbounded. Our result indicates that the removal of spatial extension and local density limit has made unbounded growth of higher-order entity size more natural and more manifested in the simulations. 

\begin{table*}
\centering
\caption{Summary of curve fitting of two different growth models to the results shown in Fig.\ \ref{fig:higher-order}}
\begin{tabular}{llll}
\hline
\multicolumn{1}{c}{Data} & Model & \multicolumn{2}{c}{Results} \\
\hline
Maximum number of individual entities & Bounded growth &  \multicolumn{2}{l}{Best fit:} \\
in replicating multisets (Fig.\ \ref{fig:higher-order} top) & $n(t) = -a / \log t + b$ & 
\multicolumn{2}{l}{$n(t) = -61.7165 / \log t + 13.7754$} \\
 & & $R^2$ & $0.994446$\\
 & & AIC & $1392.47$ \\
 & & BIC & $1409.12$ \\
 \cline{2-4}
 & Unbounded growth &  \multicolumn{2}{l}{Best fit:} \\
 & $n(t) = a \log t + b$ & 
\multicolumn{2}{l}{$n(t) = 1.60238 \log t - 6.30145$} \\
 & & $R^2$ & $0.997065$\\
 & & AIC & $180.662$ \\
 & & BIC & $197.310$ \\
\hline
Average number of individual entities & Bounded growth &  \multicolumn{2}{l}{Best fit:} \\
in replicating multisets (Fig.\ \ref{fig:higher-order} bottom) & $n(t) = -a / \log t + b$ & 
\multicolumn{2}{l}{$n(t) = -57.1218 / \log t + 12.7686$} \\
 & & $R^2$ & $0.995616$\\
 & & AIC & $662.749$ \\
 & & BIC & $679.398$ \\
 \cline{2-4}
 & Unbounded growth &  \multicolumn{2}{l}{Best fit:} \\
 & $n(t) = a \log t + b$ & 
\multicolumn{2}{l}{$n(t) = 1.48336 \log t - 5.81543$} \\
 & & $R^2$ & $0.998238$\\
 & & AIC & $-1068.74$ \\
 & & BIC & $-1052.09$ \\ 
\hline
\end{tabular}
\label{tab:curvefits}
\end{table*}

\begin{figure}
\centering
\includegraphics[width=\columnwidth]{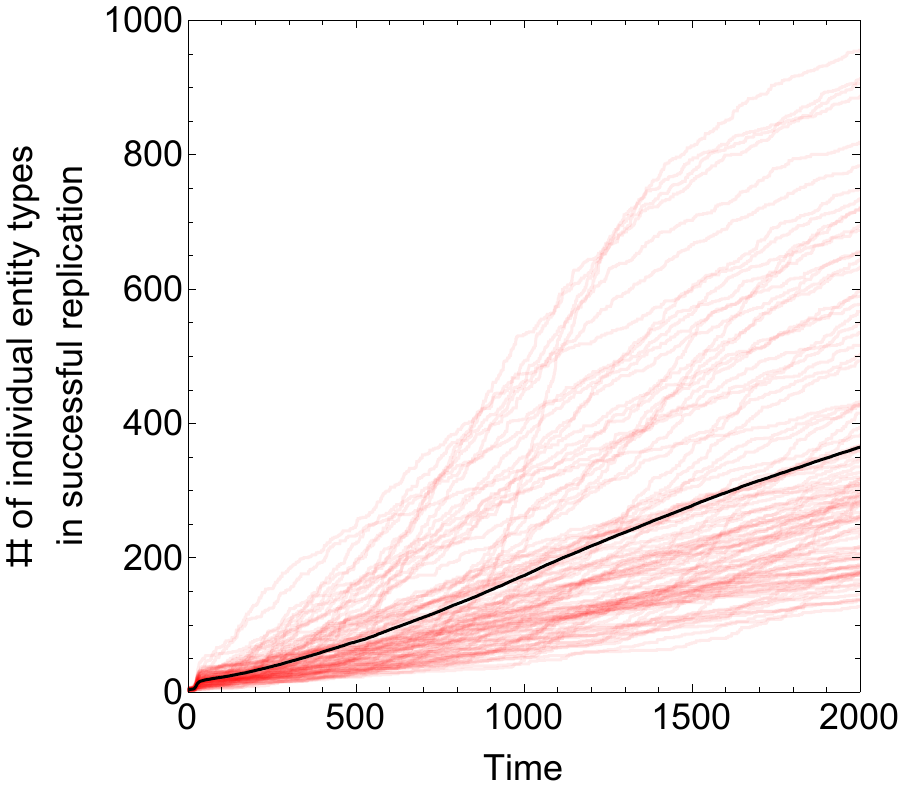}
\includegraphics[width=\columnwidth]{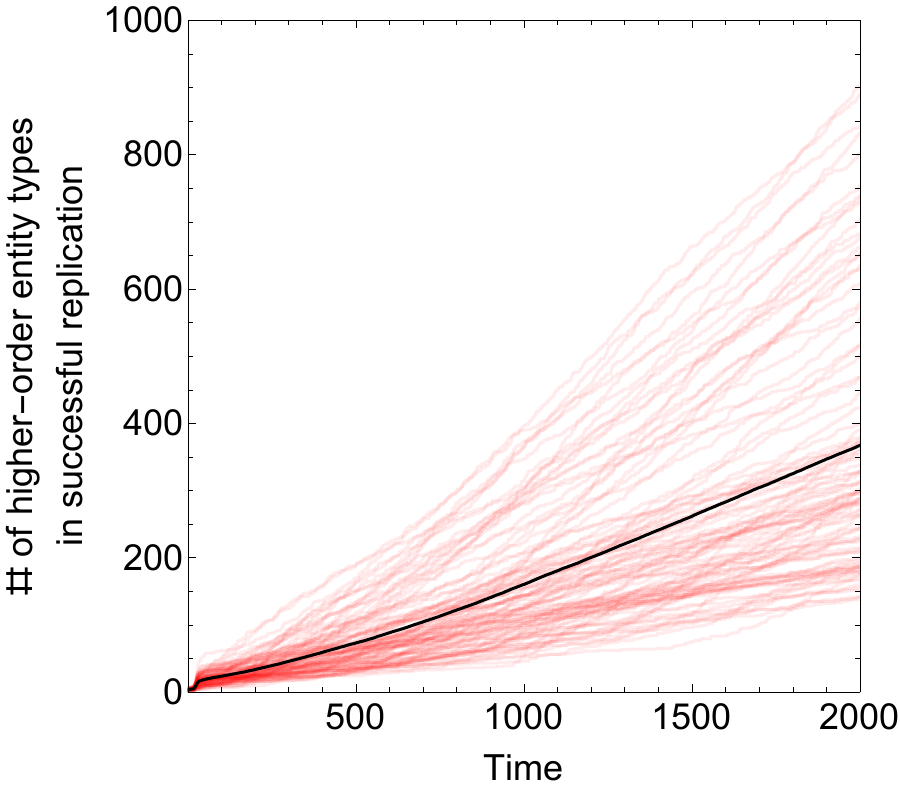}
\caption{Cumulative numbers of unique entity types that appeared in successful replication in the course of simulation. Top: Number of individual entity types. Bottom: Number of higher-order entity (multiset) types. The red curves show results of 100 independent simulation runs, while the black solid curve shows their average. The time is in linear scale (only in this figure) to show the difference in the shapes of the curves more clearly. }
\label{fig:open-ended}
\end{figure}

Finally, Figure \ref{fig:open-ended} shows the cumulative numbers of unique entity types that appeared in successful replication in the course of simulation run (top: individual entity types; bottom: higher-order entity (multiset) types). The number of higher-order entity types (Fig.\ \ref{fig:open-ended} bottom) appeared to exhibit more unbounded growth-like behavior than the number of individual entity types (Fig.\ \ref{fig:open-ended} top). However, the numbers were much lower than those reported in \cite{sayama2019}\footnote{In the original model \cite{sayama2019}, the number of unique individual entity types quickly reached the maximum value 1,000, and the number of unique higher-order entity types kept increasing to tens of thousands without bound.}, and the quantitative difference between the numbers of individual and higher-order types was not as salient as in \cite{sayama2019}. These differences can be understood in that, in the non-spatial model, evolutionary competition and selection occurs globally in a well-mixed population and thus significantly suppresses diversification of evolving entities. This fact should be taken into account as a potential drawback in using non-spatial evolutionary systems, especially if such a system is built for exploration of a vast possibility space, for which spatial models may be more suitable.

\section{Conclusions}

In this paper, we developed a simplified non-spatial version of the Hash Chemistry artificial chemistry model. The new model eliminated the assumption of a two-dimensional spatial domain and represented the spatial couplings among individual entities explicitly in the form of multisets of those entities. Each multiset essentially acted as a ``molecule'', i.e., the basic unit of replication, competition and selection. In addition, the mechanisms for competition and selection were also modeled as a simple repetition of pairwise matches between two randomly chosen multisets. These revisions were an implementation of protection and preservation of higher-order structures in evolution, which was in fact suggested as one of possible future directions in \cite{sayama2019}.

The above model simplification made numerical simulations more than two times faster than those with the original model. This allowed us to conduct a large number of independent simulation runs within a reasonable amount of time. The results of numerical simulations showed much more significant unbounded growth in both maximal and average sizes of successfully replicating higher-order entities than the original model. This result clearly demonstrates the effectiveness of the proposed non-spatial model as a minimalistic example of open-ended evolutionary systems. It also serves as concrete evidence that spatial extension is not a necessary ingredient for open-endedness in evolutionary systems. Meanwhile, it was also found that the diversities of replicating entities (either individual or higher-order ones) were substantially lower in the non-spatial model than in the original one. This indicates that, while not essential for open-endedness, spatial extension still has strong positive effects on diversification of evolving entities.

Another interesting point we note is that, in the proposed simulation algorithm (see Section \ref{algorithm}), replication of a multiset was simulated as {\em exact} replication with no mutation involved while mutations were applied to 1\% of all the multisets at a later stage within each iteration. These model settings were intentionally made so that large multisets containing many individual entities could still replicate accurately without errors. We actually tested other versions of the model in which point-wise mutations were applied to a copy of the replicating multiset in every replication event, but those models did not exhibit complexity growth because multisets would eventually become unable to accurately replicate themselves due to errors by mutation when they became too large (results not shown). This highlights the importance of high fidelity of replication (and active error correction) for complexity growth of evolving entities.

Finally, we note that some important nontrivial features that were present in the original model, {\em context dependence of fitness} and {\em multiscale adaptation}, are lost in the proposed non-spatial Hash Chemistry model. In the original model, higher-order entities were constructed each time from random samples of spatially proximate individual entities without any preserved ``molecular'' structures. Therefore, evolving entities had to improve their fitness in the context of what other entities were physically nearby, and they had to produce high fitness values at various scales (i.e., different numbers and combinations of individual entities with which they are grouped) all at once for their survival. In the proposed non-spatial model, however, such interactions among evolving entities do not exist, and evolution is reduced to a mere refinement of fitness values of mutually independent multisets. In other words, the evolutionary dynamics exhibited in the non-spatial model are fundamentally simpler with no interactions among evolving entities. It remains an open question how nontrivial ecological interactions could be introduced to this non-spatial Hash Chemistry model.

\bibliographystyle{IEEEtran}
\bibliography{sayama-IEEE-WCCI2024}

\end{document}